\pdfoutput=1
\documentclass[11pt]{article}

\usepackage[]{ACL2023}

\usepackage{times}
\usepackage{latexsym}
\usepackage{amsmath}
\usepackage{algorithm}
\usepackage{graphicx}
\usepackage{amsfonts}
\usepackage{dsfont}
\usepackage{caption}
\usepackage{float} 
\usepackage{subfigure}
\usepackage{subcaption}
\usepackage{makecell}
\usepackage{hyperref}
\usepackage{multirow}
\usepackage[T1]{fontenc}
\usepackage[utf8]{inputenc}

\usepackage{microtype}

\usepackage{inconsolata}

\title{Relative Counterfactual Contrastive Learning for Mitigating Pretrained Stance Bias in Stance Detection}

\author{Jiarui Zhang \and Shaojuan Wu \and Xiaowang Zhang \and Zhiyong Feng \\
        Tianjin University \\ \texttt{\{jiaruizhang,shaojuanwu,xiaowangzhang,zyfeng\}@tju.edu.cn}}
\begin{document}
\maketitle
\begin{abstract}
Stance detection classifies stance relations (namely, Favor, Against, or Neither) between comments and targets. Pretrained language models (PLMs) are widely used to mine the stance relation to improve the performance of stance detection through pretrained knowledge. However, PLMs also embed ``bad'' pretrained knowledge concerning stance into the extracted stance relation semantics, resulting in pretrained stance bias. It is not trivial to measure pretrained stance bias due to its weak quantifiability. In this paper, we propose Relative Counterfactual Contrastive Learning (RCCL), in which pretrained stance bias is mitigated as relative stance bias instead of absolute stance bias to overtake the difficulty of measuring bias. Firstly, we present a new structural causal model for characterizing complicated relationships among context, PLMs and stance relations to locate pretrained stance bias. Then, based on masked language model prediction, we present a target-aware relative stance sample generation method for obtaining relative bias. Finally, we use contrastive learning based on counterfactual theory to mitigate pretrained stance bias and preserve context stance relation. Experiments show that the proposed method is superior to stance detection and debiasing baselines.

\end{abstract}

\section{Introduction}

% \begin{figure}[t]
%     \centering
%     \subfigure[The examples of stance detection]{
%         \includegraphics[width=0.4\textwidth]{fig/example3.pdf}
%     }
%     \subfigure[The stance distribution of different pretrained model]{
%         \includegraphics[width=0.4\textwidth]{fig/fig1.png}
%     }
%     \caption{}
%     \label{fig:fig_micPerMon}
% \end{figure}

\begin{figure}[t]
  \centering
  \includegraphics[width=0.7\linewidth]{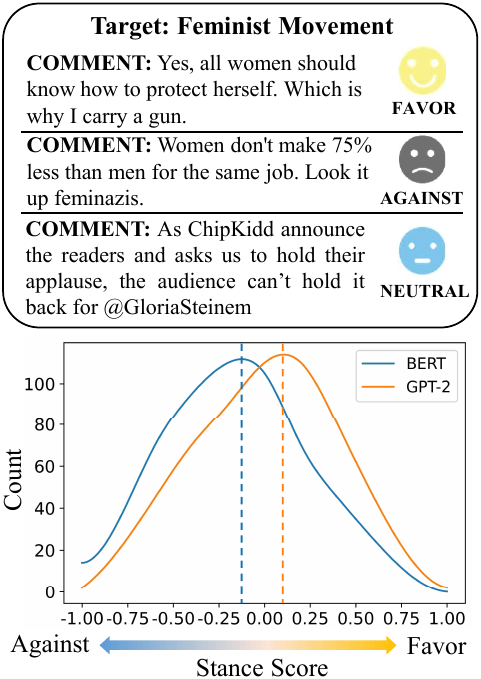}
  \caption{The examples of stance detection datasets. The stance distribution of BERT and GPT-2 for the same target ``Feminist Movement" is opposite, which reveals pretrained stance bias.}
  \label{fig1} %%% to explain more
\end{figure}

Stance is the relation between comments and targets that are explicitly mentioned or implied in the comments. Targets are entities, concepts, events, ideas, opinions, prostates, topics, etc. Stance detection aims to classify stance relation into three classes: \{Favor, Against, Neither\}\cite{mohammad-etal-2016-semeval,stance12,augenstein-etal-2016-stance}. A wide range of applications can benefit from stance detection, e.g., veracity and rumor detection~\cite{10.1145/3295823,hardalov-etal-2022-survey} debunks rumors by cross-checking the stances conveyed by the text of relevant comments.

% Fake news detection and automatic fact checking\cite{10.1145/3369026} track the spread and travel trajectory of false news based on the stance, and further investigate specific fake news by constructing the stance evolution model.Therefore, the stance relation is concealed and hidden

The comments may not directly mention whether the stance on the target is Favor or Against. Hence, stance detection models need to understand the stance relation semantics as well as the context semantics. It becomes difficult to extract stance relation semantics hidden in context semantics.
Existing works extract such implicit relation through Pretrained Language Models (PLMs) taking advantage of pretrained knowledge in learning the specific task semantics~\cite{zhang-etal-2020-enhancing-cross,allaway-2020,ji2023counterfactual,2021pretrained}. PLMs use as much training data as possible to extract the common features of the context. They embed the pretrained knowledge into the stance relation semantics as common features and then improve the performance. However, training data often contains specific stances. As a result, the stance implied in pretrained knowledge inevitably changes the original stance of samples. It causes \emph{pretrained stance bias}, as a kind of bias in stance detection~\cite{feng-etal-2023-pretraining}, which interferes with detecting ``\emph{real} stance''.
As shown in Figure~\ref{fig1}, for the same target, "Feminist Movement", the distribution tends to be against for BERT and favor for GPT-2.

%the stance distribution obtained by BERT and GPT-2 is different.
%This stance deviation causes the stance classification of the model to be uncontrollable in real application, which should be used with caution, otherwise moral problems will occur. Palomino et al.\cite{palomino-etal-2022-differential} showed that differences in stance bias are perceptible by training or visualization methods. Yuan et al. \cite{yuan2022debiasing} subtracted the direct text effect from the total causal effect to reduce the dataset bias in the text part. Wu et al. \cite{10.1007/978-3-031-26390-3_17} proposed Publisher Style Aggregation, a generalizable approach aggregating publisher posting records to learn writing style and veracity stance. 

% PLMs have political leanings which reinforce the polarization present in pretraining corpora propagating social biases\cite{feng-etal-2023-pretraining}. And the differences in stance are perceptible by training or visualization methods\cite{palomino-etal-2022-differential}. 

Recently, there are some works about bias in stance detection.
Some researches find that the differences in stance bias are perceptible by training or visualization methods~\cite{palomino-etal-2022-differential,feng-etal-2023-pretraining}. SSR~\cite{yuan2022debiasing} incorporates the stance reasoning process as task knowledge to assist in learning genuine features and reducing dataset bias. PSA~\cite{10.1007/978-3-031-26390-3_17} proposes a generalizable approach aggregating publisher posting records to learn veracity stance.
%In a short, existing approaches have studied to reduce the stance bias caused by training data. 
%
However, existing approaches have less investigated to measure the pretrained stance bias brought by PLMs.
Pretrained stance bias is weak quantifiability since absolute measurement requests a central point (taken as \emph{anchor}) in stance semantic space. 
%different PLMs have different potential stance tendencies~\cite{feng-etal-2023-pretraining}. 

In this paper, we propose Relative Counterfactual Contrastive Learning (RCCL) to measure and mitigate pretrained stance bias. Instead of absolute measurement, we turn to relative measurement via itself as an anchor for overtaking its weak quantifiability.
Firstly, we apply the causal model to analyze the relationship between the pretrained model, the context, and the stance. It enables us to capture the pretrained stance bias as the direct causal effect of the text on the stance. Then, we employ a target-aware relative stance sample generation method, creating counterfactual samples that have a relative stance to the original samples. Finally, we use counterfactual contrastive learning to drive stance relation representation to change in the context stance space rather than the pretrained stance space. We use positive and negative samples as relative anchors, mitigating the pretrained stance bias while preserving the original context stance relation. Experiments show that our proposed method outperforms existing debiasing and stance detection baselines.

%The main contributions of this paper can be summarized as follows: %%% to be improved
%\begin{itemize}
% \item We first construct a Structural Causal Model (SCM) to formalize the causalities to alleviate pretrained stance biases. The SCM indicates that the pretrained knowledge is inherently a confounder that can lead to spurious correlations between context representations of comments and stance labels.
%   \item We propose an effective implementation to intervene in stance detection based on the SCM. We develop effective counterfactual contrastive learning to constrain the confounder, which eliminates the pretrained stance knowledge and relies more on robust stance features in specific samples during the training and testing phase.
 %  \item Extensive experiments on the benchmark dataset show that the proposed framework achieves state-of-the-art performance in stance detection. We also extend the approach to zero-shot and few-shot stance detection and related challenges to demonstrate the superiority and generalization of our approach.
 %\end{itemize}

\section{Related Works}
In this section, we discuss related works in two categories: stance detection and debiasing strategy.
\subsection{Stance Detection}
Stance detection aims to identify the attitude from a comment towards a certain target. Most early stance detection research focused on online English forums and political debates. In 2016, Related shared tasks at conferences such as SemEval-2016 \cite{mohammad-etal-2016-semeval} and Catalan tweets \cite{Taul2017OverviewOT} have led to a surge in the number of studies on stance detection. 

% Thomas et al.\cite{thomas-etal-2006-get} collected the records of speeches and debates in the US Congress in 2005, and marked the support and opposition stances. 
The earlier work only trained the model and made predictions on a single target \cite{augenstein-etal-2016-stance}. Stancy \cite{popat-etal-2019-stancy} leveraged BERT representations learned over massive external corpora and utilized consistency constraints to model target and text jointly. S-MDMT\cite{wang2021solving} applied the target adversarial learning to capture stance-related features shared by all targets and combined target descriptors for learning stance-informative features correlating to specific targets. TGA-Net\cite{allaway-2020} used learned scaled dot product attention to capture the relation between the target and other related targets and comments.

%Liu et al.\cite{9746739} establish the connections between targets and enhance the model's generalization ability. Other works introduce external knowledge\cite{10.1145/3293318,10.1145/3386252}. Zhang et al.\cite{zhang-etal-2020-enhancing-cross} mapped words to the semantic and emotional lexicon.

Unlike the above work, we focus on pretrained stance bias. To our knowledge, we are the first to introduce counterfactual reasoning to stance detection.

\subsection{Debiasing Strategy}
Recent debiasing techniques integrate the idea of counterfactual reasoning into their frameworks across multiple tasks such as question answering\cite{NEURIPS2021_878d5691}, visual question answering \cite{Niu_2021_CVPR}, text classification\cite{qian-etal-2021-counterfactual}, and recommendation \cite{ji2023counterfactual}. Recently, causal inference has also attracted increasing attention in natural language processing to mitigate the dataset bias \cite{udomcharoenchaikit-etal-2022-mitigating}.

As a common debiasing representation learning method, contrastive learning uses comparative learning to capture the subtle features of data. Recent research reveals some key considerations for better contrastive learning, such as heavy data augmentation \cite{lopes2020affinity} and large sets of negatives \cite{chen2020simple}. Some work also introduces counterfactual and causal ideas. Among the existing counterfactual contrastive learning methods, C2l\cite{choi2022c2l} is similar to ours. It synthesizes “a set” of counterfactuals and makes a collective decision on the distribution of predictions on this set, which can robustly supervise the causality of each term. We also propose a novel contrastive learning framework guided by causal inference for mitigating pretrained stance bias. Our negative samples and positive samples are generated from the original samples.

\section{Methods}
\subsection{Problem Statement}
Given a comment $c$, and a target $t$, let $D = \{x_i = (c_i, t_i, y_i)\}^N_{i=1}$ be a stance detection dataset with $N$ examples, $y_i$ is a stance label.  The goal is to predict a stance label $\hat{y} \in \{Favor, Againist, Neutral\}$ for each $c_i$ and $t_i$. The context representation of $c_i$ and $t_i$ is $x$, encoded by the pretrained model $T$. $T$ and a classifier $P(y|x;\theta)$ are fine-tuned for stance detection on the training dataset and then evaluated on the testing dataset.

% Given a comment composed of $n$ words $c = \{w_1, w_2, ..., w_n\}$, and a target composed of $m$ words $t = \{a_1, a_2, ..., a_m\}$.

\begin{figure}[t]
	\centering
	\begin{minipage}{0.46\linewidth}
		\centering
		\includegraphics[width=0.9\linewidth]{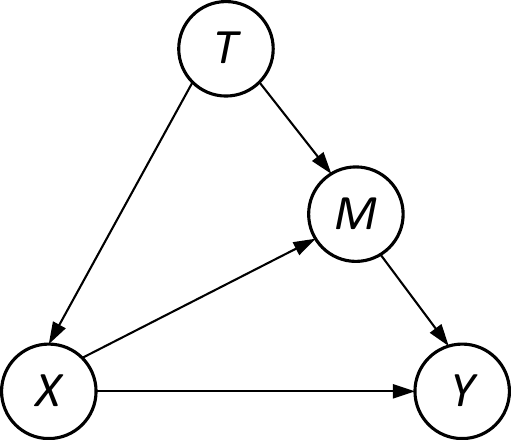}
		\label{chutian1}%文中引用该图片代号
	\end{minipage}
	%\qquad
	\begin{minipage}{0.49\linewidth}
		\centering
		\includegraphics[width=0.9\linewidth]{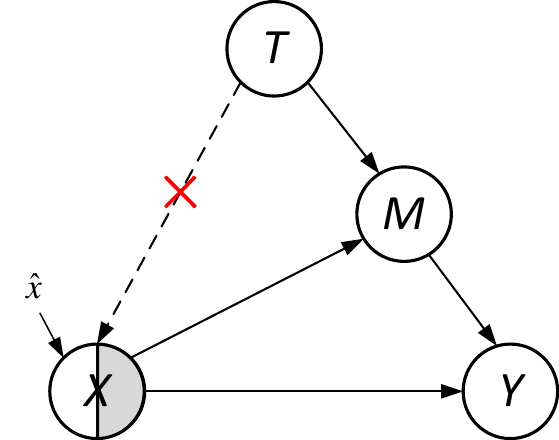}
		\label{chutian2}%文中引用该图片代号
	\end{minipage}
        \caption{(a) Causal graph for stance detection. (b) Interventional stance detection where we directly model $P(Y|do(X=\hat{x}))$.}
        \label{fig2}
\end{figure}

\subsection{Structural Causal Model for Stance Detection}

SCM is represented as a directed acyclic causal graph $G = \{V, E\}$, where $V$ denotes the set of variables and $E$ represents the cause-and-effect relationships. The nodes denote the variables in the model, and the edges between nodes denote the causality. For example, if $Y$ is a descendant of $X$, $X$ is a potential cause of $Y$ and $Y$ is the effect. As the previous discussion, $\theta$ in fine-tuning is dependent on the pretrained model. This ``dependency” can be formalized with a Structural Causal Model (SCM) \cite{primercausal} as shown in Figure~\ref{fig2}(a). The SCM for stance detection's detailed implementations are:

% Figure~\ref{} shows an example of causal graph that consists of three variables. If the variable $X$ has a direct effect on the variable $Y$, we say that $Y$ is the child of $X$, i.e., $X\rightarrow Y $. If X has an indirect effect on $Y$ via the variable $M$, we say that $M$ acts as a mediator between $X$ and $Y$, i.e.,$ X\rightarrow M\rightarrow Y$.

\begin{itemize}
\item {$T \rightarrow X$.}
We denote $X$ as the context representation of comments and targets and $T$ as the pretrained model. The connection means that the representation $X$ is generated by $T$.
\item {$X \rightarrow M \leftarrow T$.}
$M$ is a mediator variable that denotes the low-dimensional multi-source knowledge of comments, targets, and $T$. The branch $X \rightarrow M$ means the representation can be denoted by linear or nonlinear projection onto the manifold base. Moreover, $T \rightarrow M$ denotes the semantic information embedded in $M$. 
\item {$X \rightarrow Y \leftarrow M$.}
To simplify the description, we directly denote $Y$ as the probability of predicting stances. $X$ affects $Y$ in two ways, the direct path $X \rightarrow Y$ and the mediation path $X \rightarrow M \rightarrow Y$. 
% $X \rightarrow Y$ can be neglected if $X$ can be fully represented by $M$, which is almost impossible for a model. The mediation path is also unavoidable because any classifier is considered to utilize $M$ implicitly.
\end{itemize}

\begin{figure*}[t]
  \centering
  \includegraphics[width=\linewidth]{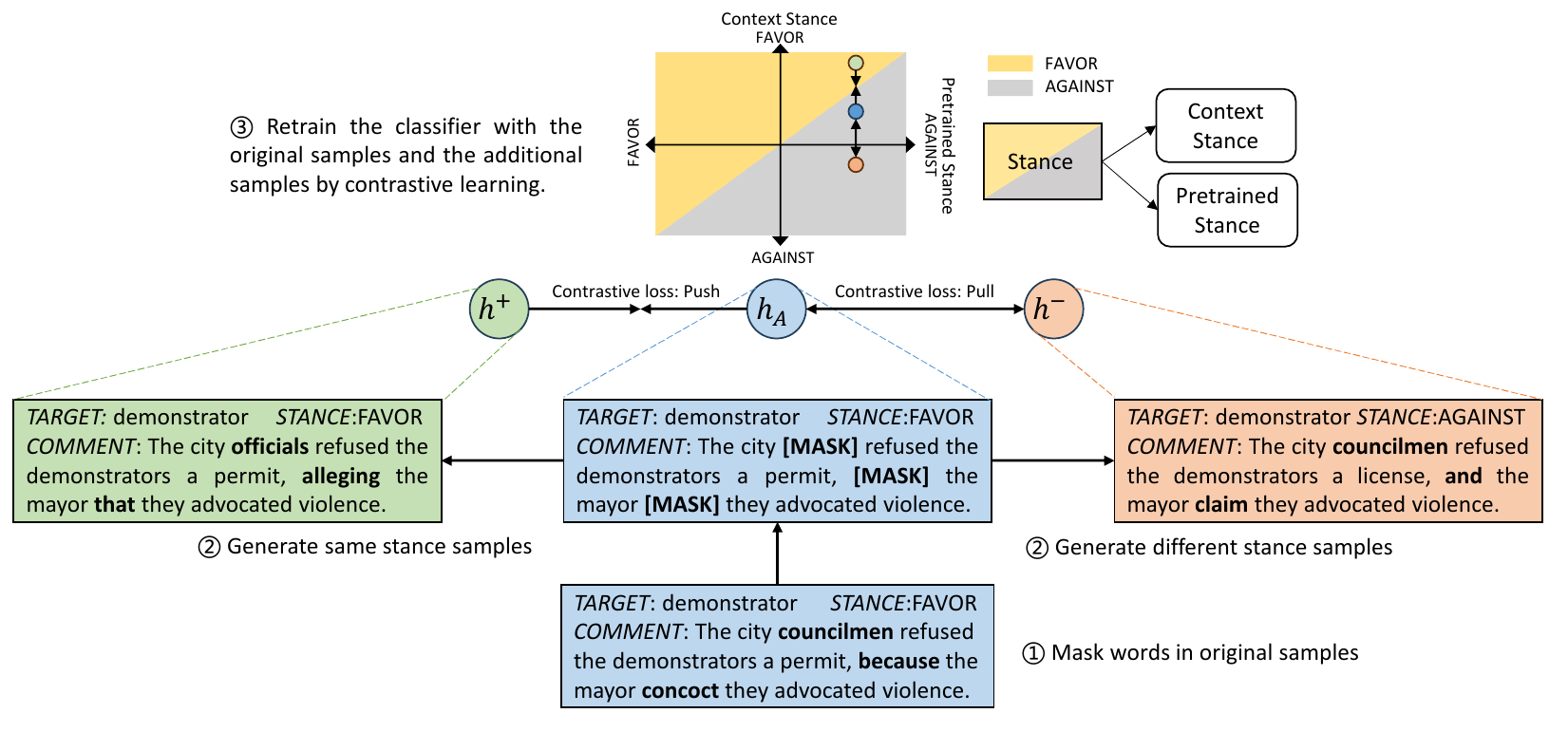}
  \caption{The overall architecture of relative counterfactual contrastive learning.}
  \label{fig3-1}
\end{figure*}

An ideal stance detection model should capture the true causality between $X$ and $Y$ to adapt to various cases. However, the traditional methods that use the correlation $P(Y|X)$ fail to do so because $X$ is not the only potential cause of $Y$. Therefore, the increased probability of $Y$ given $X$ will be affected by the spurious correlation via the two paths: $T \rightarrow X$ and $T \rightarrow M \rightarrow Y$. Accroding to causal theory, we can use the causal intervention $P(Y|do(X))$ instead of the likelihood $P(Y|X)$ for stance detection to exploit the true causality between $X$ and $Y$ as shown in Figure \ref{fig2}(b).

Then the model is inadvertently contaminated by the spurious causal correlation: $X \rightarrow M \leftarrow T$ , a.k.a. a back-door path in causal theory. To decouple the spurious causal correlation, the back-door adjustment predicts an actively intervened answer via the do(·) operation:

\begin{small}
\begin{equation}
\begin{split}
P(Y \mid d o(X=\hat{x}))=\sum_k P(Y \mid X=\hat{x}, T=t) P(T=t)& \\
\quad=\sum_k P(Y \mid X=\hat{x}, T=t, M=g(x, t)) P(T=t)&
\end{split}
\end{equation}
\end{small}
where $\hat{x}$ could be any counterfactual embedding as long as it is no longer dependent on $T$ to detach the connection between $X$ and $T$. 
%In the light of the social science’s view, humans rely on exploring contrastive attributes between the objective factual and their potential counterfactuals for classification\cite{MILLER20191}. Therefore, we will construct counterfactual samples for $X$ to detach the spurious connection between the contexts and labels.

% \begin{equation}
% \begin{aligned}
% & f(\tilde{x})=f\left(\left\langle w_1, w_2, \cdots, w_n\right\rangle\right) \\
% & \forall w_i \in \tilde{x}, \begin{cases}w_i \leftarrow[\text { MASK] } & \text { if } w_i \in x_{\text {content }} \\
% w_i \leftarrow w_i & \text { if } w_i \in x_{\text {context }}\end{cases}
% \end{aligned}
% \end{equation}

% naturally reflects as the keyword bias captured by M for a specific text document x, where $x_{\text {content }}$ and $x_{\text {context }}$ denote the main content and the context of x, respectively.
% \subsection{Base Classification Model}

% Formally, given a training set $\{x_i = (c_i, t_i, y_i)\}^N_{i=1}$ where $c_i$ is comment (possibly a set of text pieces or a single piece), $t_i$ is target, and $yi \in Y$ is a label. The text classifier $f: c,t \rightarrow y$ maps an input $(c_i, t_i)$ to the corresponding label $\hat{{y}}_i$. It is trained to minimize the cross-entropy loss $\mathcal{L}_{CLS}$ between the predicted label $\hat{{y}}_i$ and the ground-truth label $y$:
% \begin{equation}
% \mathcal{L}_{CLS}=-\frac{1}{N}\sum_{i=1}^{N} y_i \log \hat{y}_i
% \end{equation}

\subsection{Relative Stance Samples Generation}

The causal intervention operation wipes out the incoming links of a cause variable $X$, encouraging the model $T$ to infer. We attempt to utilize a partially blindfolded counterfactual sample.
%Specifically, we deliberately expose some words to the trained model to exhibit their potentially negative influence. 

We first train a classifier $f$ (e.g., fine-tuning a pretrained model) with stance detection dataset. Then, we generate relative stance samples by masking a fixed percentage of the input tokens, which is replaced by a special tag [MASK]. However, comments may contain targets or related words that directly refer to targets, and removing those tokens will make a big difference to the sample. To avoid this, we will skip the target-related keywords while masking tokens. Inspired by the recent counterfactual sample study \cite{feder2021causalm}, we use NLTK\footnote{https://www.nltk.org/} to extract target-related words that may destroy the sample. We only mask target-unrelated words. For each samples $x_i$, we generate a set of samples $S_i=\left\{\tilde{x}_{i, 1}, \tilde{x}_{i, 2}, \cdots\right\}$. Then we use T5\cite{10.5555/3455716.3455856} to fill the samples in $S_i$ and obtain the new samples $\hat{S}_i=\left\{\hat{x}_{i, 1}, \hat{x}_{i, 2}, \cdots\right\}$. Even if samples in $\hat{S}_i$ share similar syntactic structures and words, the stance labels cannot be determined, which are counterfactual samples.

We use the classifier $f$ to classify the stance label $\hat{y}_{i}$ of generated samples:
\begin{equation}
\hat{y}_{i}=\underset{y}{\arg \max } \hat{p}(y \mid \hat{x}_{i})
\end{equation}
% Empirically, the average ratio of contents to contexts produced by Jieba on all datasets is approximately 87\%:13\%. 
% Therefore, the mask strategy is:
% \begin{equation}
% \forall w_i \in {c}, \begin{cases}w_i \leftarrow[\text { MASK }] & \text { if } w_i \in \{w_m\} \\ w_i \leftarrow w_i & \text { if } w_i \in \{w_c\}\end{cases}
% \end{equation}
% Then, we use the classifier $f$ to select generated samples with the largest probabilities for each label. We denote the replaced instance as $\hat{S}_i=\left\{\hat{x}_{i, 1}, \hat{x}_{i, 2}, \cdots\right\}$. 

The final stance relation consists of context stance and pretrained stance. We mask part of the context and generate counterfactual samples with different stance labels, which makes the model more sensitive to context stance. Finally, we retrain the classifier with the original samples and the additional samples.

%首先，我们

% We first perform context-based negative sampling to find semantically most similar, but labeled differently, samples based on text similarity scores. Use this sample and the original sample to synthesize counterfactual samples. 

% Since not all negative samples can serve as useful negative samples, indiscriminate use of such data can compromise model robustness and algorithmic efficiency. Therefore, during the training process we only use the most semantically similar text features to generate counterfactual text features in each batch.

% Specifically, we first measure the text similarity between instances using the BERTScore, which uses BERT context embedding to calculate the paired cosine similarity between the reference and candidate sentences\cite{Zhang2020BERTScore}. We compute a similar matrix with the values of each element:

% \begin{equation}
% \operatorname{sim}(i, j)=\text { BERTScore }\left(\boldsymbol{h}_i, \boldsymbol{h}_j\right)
% \end{equation}

% Then, we use the regional sample $\boldsymbol{h}_{d}$ and the most similar sample $\boldsymbol{h}_{s}$ as the negative sample to construct counterfactuals:

% \begin{equation}
% \boldsymbol{h}^{-}_{d}=(1-\alpha) \circ \boldsymbol{h}_{d}+\alpha \circ \boldsymbol{h}_{s}
% \end{equation}
% where $\circ$ is the element-wise multiplication and $\alpha$ is the parameter controlling the amount of negative feature that replaces the positive feature.

\begin{table*}[]
\centering
\caption{Data split statistics for SemEval-2016 Task 6 Sub-task A.}
\begin{tabular}{lcccccccc}
\hline
Target & \#Train & Favor & Against & Neutral & \#Test & Favor & Against & Neutral \\ \hline
AT     & 513     & 92    & 304     & 117     & 220    & 32    & 160     & 28      \\
CC     & 395     & 212   & 15      & 168     & 169    & 123   & 11      & 35      \\
FM     & 664     & 210   & 328     & 126     & 285    & 58    & 183     & 44      \\
HC     & 689     & 118   & 393     & 178     & 295    & 45    & 172     & 78      \\
LA     & 653     & 121   & 355     & 177     & 280    & 46    & 189     & 45      \\ \hline
Total  & 2914    & 753   & 1395    & 766     & 1249   & 304   & 715     & 230     \\ \hline
\end{tabular}
\label{tab:sem-data}
\end{table*}

% \begin{table}[]
% \centering
% \caption{Data split statistics for SemEval-2016 Task 6 Sub-task A.}
% \begin{tabular}{lcccc}
% \hline
% Target  & Favor & Against & Neutral & Total \\ \hline
% AT     & 124    & 464    & 277     & 713    \\
% CC     & 425     & 212   & 15      & 564    \\
% FM     & 664     & 210   & 328     & 126      \\
% HC     & 689     & 118   & 393     & 178      \\
% LA     & 653     & 121   & 355     & 177      \\ \hline
% Total  & 2914    & 753   & 1395    & 766     \\ \hline
% \end{tabular}
% \label{tab:sem}
% \end{table}

% \begin{table}[]
% \caption{Data split statistics for UKP.}
% \begin{tabular}{lcccc}
% \hline
% Target & Favor & Against & Neutral & Total \\ \hline
% AB     & 680   & 822     & 2427    & 3929  \\
% CL     & 706   & 839     & 1494    & 3039  \\
% DP     & 457   & 1111    & 2083    & 3651  \\
% GC     & 787   & 665     & 1889    & 3341  \\
% ML     & 587   & 636     & 1262    & 2475  \\
% MW     & 576   & 551     & 1346    & 2473  \\
% NE     & 606   & 852     & 2118    & 3576  \\
% SU     & 545   & 729     & 1734    & 3008  \\ \hline
% Total  & 4944  & 6195    & 14353   & 25492 \\ \hline
% \end{tabular}
% \label{tab:ukp}
% \end{table}

\subsection{Counterfactual Contrastive Learning}

The overall architecture of our model is shown in Figure \ref{fig3-1}. For each sample $x_i$, we generate a counterfactual sample set $\hat{S}_i=\left\{\hat{x}_{i, 1}, \hat{x}_{i, 2}, \cdots\right\}$ where $\hat{x}_{i}=(\hat{c}_{i}, \hat{t}_{i}, \hat{y}_{i})$ in Section 3.3. By learning to contrast
with these samples, it is beneficial for the uniformity of stance relation representations. We denote the stance-same samples as positive samples, $\hat{S}_i^+=\left\{\hat{x}_{i, 1}^+, \hat{x}_{i, 2}^+, \cdots\right\}$, while stance-different samples are negative samples $\hat{S}_i^-=\left\{\hat{x}_{i, 1}^-, \hat{x}_{i, 2}^-, \cdots\right\}$. Then we use BERT to get their representations $h_i$, $\{h_{i,p}^+\}^P_{p=1}$ and $\{h_{i,q}^-\}^Q_{q=1}$, respectively. We adopt the following margin-based ranking loss for model training:
\begin{equation}
\begin{aligned}
& \mathcal{L}_{CONTRA}=-\frac{1}{N}\sum_{i=1}^{N} \max (0, \\
& \Delta_m+\frac{1}{P} \sum_{p=1}^P s_\theta\left(h_i, h_{i,p}^{+}\right)-\frac{1}{Q} \sum_{q=1}^Q s_\theta\left(h_i, h_{i,q}^{-}\right)),
\end{aligned}
\end{equation}
where $P,Q$ is the number of positive and negative samples, $\Delta_m$ is a margin value which we set to 1 in this work, and $s_\theta(·, ·)$ denotes the distance between the BERT representations.

\subsection{Model Training}

Formally, given a training set $\{x_i = (c_i, t_i, y_i)\}^N_{i=1}$ where $c_i$ is comment (possibly a set of text pieces or a single piece), $t_i$ is target, and $y_i \in Y$ is a label. The text classifier $f: c,t \rightarrow y$ maps an input $(c_i, t_i)$ to the corresponding label $\hat{{y}}_i$. It is trained to minimize the cross-entropy loss $\mathcal{L}_{CLS}$ between the predicted label $\hat{{y}}_i$ and the ground-truth label $y_i$:
\begin{equation}
\mathcal{L}_{CLS}=-\frac{1}{N}\sum_{i=1}^{N} y_i \log \hat{y}_i
\end{equation}

Finally, the parameters of the model are trained to minimize the both loss together as follows:
\begin{equation}
\mathcal{L}=\mathcal{L}_{CLS}+\gamma \mathcal{L}_{CONTRA}
\end{equation}
where $\gamma$ are hyperparameters.

\section{Experiments}
% Our experiments try to investigate the following research questions:
% \begin{itemize}
%     \item \textbf{RQ1:} Can RCCL mitigate the pretrained stance bias?
%     \item \textbf{RQ2:} Can RCCL benefit stance detection?
% \end{itemize}

% To achieve this goal, we conduct stance detection in three datasets including general datasets, zero-shot and few-shot datesets.

% In Section 4.1, we first introduce the settings of experiments. Then, we approximately demonstrate the comparison results of RCCL in Section 4.2. Section 4.3 shows an ablation study of our proposed model. Afterward, we analyze the impact of the number of generated samples and mask ratio (in Section 4.4). Finally, we focus on stance detection in the few-shot and zero-shot scenarios in Section 4.5.

\subsection{Experimental Setup}
\paragraph{Datasets.}
In our experiments, we select three datasets in stance detection task: SemEval-2016 Task 6 Sub-task A dataset\cite{mohammad-etal-2016-semeval}, UKP\cite{stab-etal-2018-cross}, and VAST\cite{allaway-2020} (mainly for few-shot and zero-shot experiments). Their data distribution are shown in Table~\ref{tab:sem-data}, Table~\ref{tab:ukp} and Table~\ref{tab:stati_vast}. %%% 选择理由
\begin{itemize}
    \item The SemEval-2016 Task 6 Sub-task A dataset contains five targets as follows: “Atheism (AT)”, “Climate Change is a real Concern (CC)”, “Feminist Movement (FM)”, “Hillary Clinton (HC)” and “Legalization of Abortion (LA)”. It is widely used for stance detection tasks.
    \item The UKP dataset consists of eight different topics, including “Abortion (AB)”, “Cloning (CL)”, “Death Penalty(DP)”, “Gun Control (GC)”, “Marijuana Legalization (ML)”, “Minimum Wage (MW)”, “Nuclear Energy (NE)” and “School Uniforms (SU)”. We adopt the train, validation, and test splits provided by the original authors. The comments in UKP are longer, making classification more difficult.
    \item VAST is a zero-shot and few-shot stance detection dataset. It consists of a large range of targets covering broad themes, such as economy (e.g., `private corporation profit’), traffic (e.g., `alternative transportation’), and religion (e.g., `Mother Teresa'). The same comment contains multiple targets and has different stances for these targets in VAST. It can be used to test the robustness of RCCL in real-world situations with few data.
\end{itemize}

Following previous work, we adopt Macro-averaged $F1$ of each label to measure the testing performance of the models. 

\begin{table}[]
\caption{Data split statistics for UKP.}
\begin{tabular}{lcccc}
\hline
Target & Favor & Against & Neutral & Total \\ \hline
AB     & 680   & 822     & 2427    & 3929  \\
CL     & 706   & 839     & 1494    & 3039  \\
DP     & 457   & 1111    & 2083    & 3651  \\
GC     & 787   & 665     & 1889    & 3341  \\
ML     & 587   & 636     & 1262    & 2475  \\
MW     & 576   & 551     & 1346    & 2473  \\
NE     & 606   & 852     & 2118    & 3576  \\
SU     & 545   & 729     & 1734    & 3008  \\ \hline
Total  & 4944  & 6195    & 14353   & 25492 \\ \hline
\end{tabular}
\label{tab:ukp}
\end{table}

\begin{table*}[]
\centering
\caption{Macro-averaged $F1$ on the SemEval-2016 dataset and UKP dataset.}
\label{tab:all-result}
\scalebox{0.84}{
\begin{tabular}{lcccccc|ccccccccc}
\hline
     \multirow{2}{*}{Model}    & \multicolumn{6}{c|}{SemEval-2016}           & \multicolumn{9}{c}{UKP}                                          \\ \cline{2-16}
    & AT   & CC   & FM   & HC   & LA   & ALL & AB   & CN   & DP   & GC   & ML   & MW   & NE   & SU   & ALL \\ \hline \hline
BiLSTM   & .605 & .420 & .516 & .558 & .591 & .538     & .470     & .556     & .489     & .458     & .535     & .511     & .491     & .509     & .502         \\ 
$\text {BERT}_\text {SEP}$   & .687 & .441 & .617 & .623 & .586 & .591     & .491 & .669 & .524 & .516 & .663 & .649 & .585 & .634 & .591     \\
$\text {BERT}_\text {MEAN}$  & .694 & .525 & .592 & .646 & .663 & .624     & .538 & .686 & .542 & .511 & .652 & .664 & .578 & .578 & .600     \\
STANCY   & .699 & .537 & .617 & .647 & .634 & .626     & .518 & .675 & .567 & .502 & .654 & .662 & .683 & .629 & .599     \\
S-MDMT     & .695 & .525 & .638 & .672 & .672 & .640     & - & - & - & - & - & - & - & - & -     \\
TAPD     & \textbf{.739} & .593 & .639 & \textbf{.700} & .672 & .669     & .549 & .719 & .572 & .517 & .670 & .699 & .590 & .641 & .620     \\ \hline
%PoE    & .637 & .545 & .603 & .642 & .632 & .619  & .531     & .663     & .568     & .545     &.674      & .689     & .531     & .658    & .608      \\
CL       & .673 & .484 & .668 & .621 & .619 & .613     & .533     & .690     & .552     & .509     &.662      & .671     & .575     & .600     &      .601    \\
$\text {C}^2\text {L}$      & .698 & .603 & .631 & .691 & .643 & .655     & .563     & .701     & .560     & .554     & .699     & .657     & .608     & .633     & .621         \\ 
SSR    & .632  & .569     & .597  & .627  & .573  & .600           & - & - & - & - & - & - & - & - & - \\ \hline
% $\text {BERT}_\text {MEAN}$  & .694 & .525 & .592 & .646 & .663 & .624     & .538 & .686 & .542 & .511 & .652 & .664 & .578 & .578 & .600     \\
RCCL(ours)     & .715 & \textbf{.605} & \textbf{.687} & .672 & \textbf{.702} & \textbf{.683}     & \textbf{.588} & \textbf{.736} & \textbf{.592} & \textbf{.561} & \textbf{.702} & \textbf{.714} & \textbf{.628} & \textbf{.683} & \textbf{.651}     \\ 
% RoBERTa    & .670 & .526 & .566 & .590 & .640 & .598 & .565 & .713  & .561 &.536  &.676  &.684  & .610 & .590  & .622      \\
% ~~+RCCL     & .737 & .598 & .672 & .665 & .731 & .681 &.609   & .761  & .602  & .597  & .711  & .735  & .667  & .632   & .663 \\
% Flan-T5    & .722 & .729 & .645 & .798 & .657 & .710 &.635  & .752 & .679 & .658 & .726 & .769 & .701 & .708 & .691     \\
% ~~+RCCL & .735 & .711 & .688 & .762 & .702 & .719  &.673   & .770  & .675  & .669  &.703   &.765   & .710  & .724  & .706  \\ 
\hline
\end{tabular}}
\end{table*}
%[470, 550, 480, 510, 490, 530, 460, 515]

\paragraph{Baselines.}
We adopt two kinds of baselines to test the efficiency of RCCL comprehensively:
\begin{itemize}
\item \textbf{Stance Detection.} For stance detection, we use $\text {BERT}_\text {SEP}$ and $\text {BERT}_\text {MEAN}$ as the basic comparisons, which predicts the stance by appending a linear classification layer to the hidden representation of [CLS] token, or by appending a linear classification layer to the mean hidden representation over all tokens. We adopt some advanced stance detection models, including Stancy\cite{popat-etal-2019-stancy}, S-MDMT\cite{wang2021solving} and TAPD\cite{10.1145/3477495.3531979}. We also compare our model with the following methods for VAST, including two models based on BiLSTM for cross-target stance detection (Cross-Net\cite{xu-etal-2018-cross}, SEKT\cite{zhang-etal-2020-enhancing-cross}); one attention-based model (TGA-Net\cite{allaway-2020}). And a prompt learning framework for stance detection. The results of these baselines are from the original work. 
\item \textbf{Bias Removal.} For bias removal, CL\cite{khosla2020supervised} is a basic implementation of contrastive learning. We collect positive (or negative) contrastive pairs from a pool of the same (or different) class samples. $\text {C}^2\text {L}$\cite{choi2022c2l} is a causally contrastive learning method for text classification, which improves counterfactual robustness. SSR\cite{yuan-etal-2022-ssr} is a debiasing strategy for stance detection. The results for CL and $\text {C}^2\text {L}$ are collected from our experiments.
\end{itemize}

\paragraph{Implementation Details.}
We use the pretrained uncased BERT as the embedding module. The learning rate is set to 1e-5,3e-5,5e-5. The coefficient $\lambda$ is set to 1e-5. Adam is utilized as the optimizer. The batch size is 16, considering the computational resources and evaluation performance trade-offs. The temperature parameter $\tau$ is set to 0.07 for contrastive losses. $\gamma$ is set to 0.1. We apply early stopping in the training process, and the patience is 5.

%Our code is available in \url{https://github.com/anonymous}.

%Finally, Section 5.5 presents a case study.

\begin{table}[]
\caption{Ablation study on the proposed RCCL model}
\centering
\label{tab:ablation}
\begin{tabular}{llllll}
\hline
\multicolumn{4}{c}{Module} & \multirow{2}{*}{Sem16} & \multirow{2}{*}{UKP} \\ \cline{1-4}
RSSG   & IB   & SS   & CCL  &                              &                      \\ \hline
\checkmark      &      &      & \checkmark   & .683                         & .651                 \\
       & \checkmark    &      & \checkmark   & .641                         & .586                 \\
       &      & \checkmark    & \checkmark   & .669                         & .618                 \\
\checkmark      &      &      &     & .672                         & .634                 \\ \hline
\end{tabular}
\end{table}

\subsection{Overall Performance}
The overall results in Table~\ref{tab:all-result} show that our model (RCCL) outperforms all baselines and debiasing baselines on the SemEval-2016 dataset and UKP dataset as well as Table \ref{tab:vast-result} on the VAST dataset.
%% 需要层次性论述实验结果
We first evaluate that our approach contributes to the effectiveness and the robustness on unbalanced SemEval-2016 dataset. As shown in Table~\ref{tab:all-result}, SemEval-2016 is unbalanced between the Favor class and the Against class. Our RCCL outperforms all baselines and does not rely on biases in the original training data. Specifically, RCCL improves FM and LA by 8.0\% and 9.2\%, respectively, which are developed more than other targets. Both FM and LA are parts of women's rights. It suggests that when the model better understands the causal correlation between the input context and the corresponding label, it is more robust against pretrained stance bias.  Although there may be differences in the difficulty of samples corresponding to two different targets, our model achieves the best score on AT (.715) while the macro-f1 score is .605 on CC. The gap between AT and CC is narrowed. These experimental results indicate that our method effectively mitigates pretrained bias, focusing more on the stance of the context itself.

Our method achieves significant improvements on the UKP dataset, with a total score of .651. Across all eight targets, our method all obtains the highest scores. In the stance detection domain, TAPD scores .669 on the SEM dataset and .620 on the UKP dataset. Compared to TAPD, $\text {C}^2\text {L}$ scored lower on the SEM dataset and higher on the UKP dataset. Both methods outperform the base BERT model. It suggests that the UKP dataset is more susceptible to bias influence, with larger variations in learning performance due to changes in the size of the training set, indicating higher data variance.

CL and $\text {C}^2\text {L}$, as debiasing methods, enhance the model's robustness and stability. Some of the effectiveness of our method also stems from contrastive learning. Moreover, our method can be considered a form of data augmentation. We generate counterfactual samples to complete the dataset. Since we select samples with high probability and filter labels after generating samples, this effectively mitigates the imbalance between categories, contributing to the overall performance.

%我们的方法在ukp数据集上也取得了显著的进步，总分为651。在所有的8个目标上，我们的方法都拿到了最好的分数。并且在立场检测领域最好的结果，也就是TAPD，在sem数据集上分数为669，在ukp数据集上为620。相比较TAPD来说，C2L在sem数据集上分数低于他，在ukp数据集上高于他。并且这两个方法都比用bert基础模型要好。这说明ukp数据集更容易被偏见影响。同样大小的训练集的变动所导致的学习性能的变化，也就是数据的方差较大。而CL和C2L都是去偏方法，增强了模型的鲁棒性和稳定性。我们方法的一部分效果也来源于此。值得一提的是，我们的方法也可以说是一种数据增强方法。我们生成了反事实样本，补全了数据集。由于我们在生成样本之后会选择置信度大的样本并且会对样本的标签进行筛选，这有效缓解了类别之间的不平衡。这对最后的整体效果也有帮助。

Our experiments show that pretrained stance bias is removed as a relative bias rather than an absolute bias, which successfully preserves context stance relations. Our method achieves state-of-the-art performance whether compared with stance detection methods or debiasing methods.

% mitigate the pretrained stance bias while benefiting stance detection.%%% 需要再凝练（实验得出结论（实验目的）与motivate一致）

% \usepackage{multirow}

\subsection{Ablation Study}
To verify the effectiveness of the two components: Relative Stance Samples Generation(RSSG) and Counterfactual Contrastive Learning(CCL), we conduct ablation study and the results shown in Table~\ref{tab:ablation} demonstrates that they indeed improve performance. 

For the selection of positive and negative samples, we test three modules. (1) RSSG: Relative Stance Samples Generation that is introduced in this paper. (2) IB: samples with different labels within the same batch as negative samples and samples with the same label as positive samples during training. (3) SS: searching the entire dataset to find semantically similar samples and dividing them as positive or negative based on whether their labels are the same. It is implemented by faiss\footnote{https://github.com/facebookresearch/faiss}.

The performance of the model has decreased overall from the experimental results. Compared to the other methods of selecting positive and negative samples, our proposed method RSSG performs the best (.683 and .651), with SS being the second-best(.669 and .618), and IB being the least effective (.641 and .586). As suggested by \cite{robinson2021contrastive}, good negative samples should have different labels that are difficult to distinguish with anchors while their semantic representations are close. RSSG can not only exhibit semantic similarity but also contain dissimilarity in the most critical feature dimensions that need to be distinguished. It makes the model more sensitive to differences in context stance relation, contributing to better outcomes.

Meanwhile, we also test the role of counterfactual contrastive learning (CCL) in our overall model. The results indicate that if we use the generated counterfactual samples as augmentation data without any processing, the model's accuracy drops to .672 and .634. Because data augmentation reaches a bottleneck after a certain size, additional augmented data does not lead to further improvements. Contrastive learning can break the bottleneck and learn the discriminative and robust features.

% As shown in the bottom of Table~\ref{tab:ablation}, model performance has varying degrees of degradation. Notably, w/o $h_d^-$ led to a greater performance degradation at a low-lens setting than a fully supervised setting, suggesting that contrast properties can further excite knowledge of classification in PLMs. w/o $h_c^-$ shows that learning the common features of the contrasting attributes between training instances is crucial. The "w/o hc" has less performance degradation in a low-lens environment than in a fully supervised environment. We conclude that an imbalance in the distribution of training data can seriously affect performance. Also, it is worth noting that our contrast attribute construction does not consider any entity information. As a result, attributes may be too spread out in the embedded space. While w/o $h_c^-$ performed slightly worse under full supervision, w/o $h_d^-$ performed significantly worse under extreme data scarcity, suggesting that document-level contrast learning can force PLMs to focus on effective contrast properties in prompts.
\begin{table}[t]
\caption{Data split statistics for VAST.}
\centering
\label{tab:stati_vast}
\tabcolsep=3pt
\begin{tabular}{llll}
\hline
                    & Train & Dev  & Test \\ \hline
\# Examples         & 13477 & 2062 & 3006 \\
\# Unique Comments  & 1845  & 682  & 786  \\ \hline
\# Zero-shot Targets & 4003  & 383  & 600  \\
\# Few-shot Targets  & 638   & 114  & 159  \\ \hline
\end{tabular}
\end{table}

\begin{figure}[t]
  \centering
  \includegraphics[width=\linewidth]{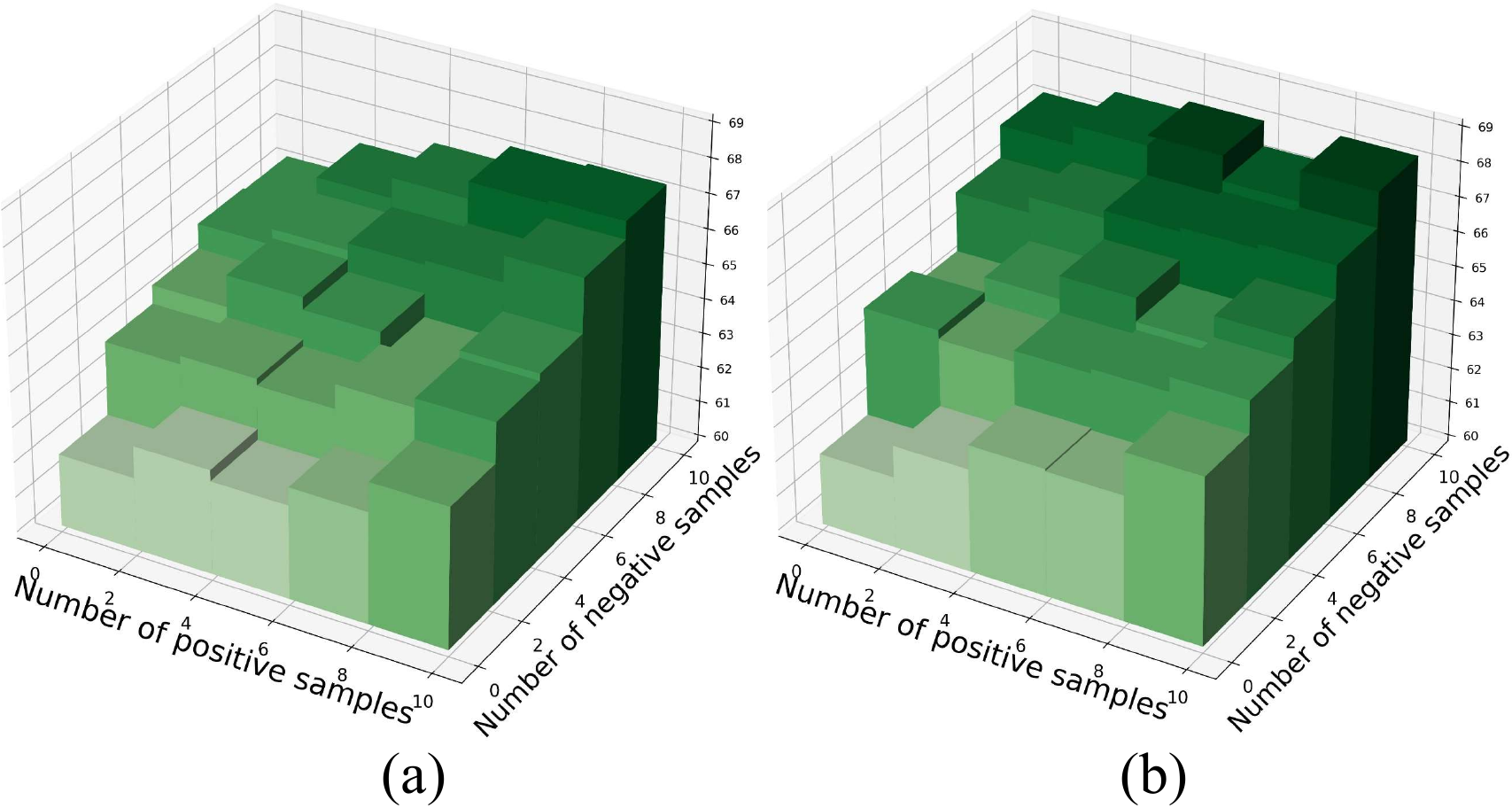}
  \caption{(a) The macro-f1 with data augmentation. (b) The macro-f1 with contrastive learning.}
  \label{fig4-1}
\end{figure}

% \begin{figure}[t]
%     \centering
%     \subfigure[1]{
%         \includegraphics[width=0.2\textwidth]{fig/fig4-2-2.pdf}
%     }
%     \subfigure[2]{
%         \includegraphics[width=0.2\textwidth]{fig/fig4-2-1.pdf}
%     }
%     \caption{1}
%     \label{fig4}
% \end{figure}

% \begin{figure}[t]
% 	\centering
% 	\begin{minipage}{0.49\linewidth}
% 		\centering
% 		\includegraphics[width=0.9\linewidth]{fig/fig5-1-1.png}
% 		\label{chutian1}%文中引用该图片代号
% 	\end{minipage}
% 	%\qquad
% 	\begin{minipage}{0.49\linewidth}
% 		\centering
% 		\includegraphics[width=0.9\linewidth]{fig/fig5-1-2.png}
% 		\label{chutian2}%文中引用该图片代号
% 	\end{minipage}
%         \caption{(a) Causal graph for stance detection. (b) Interventional stance detection where we directly model $P(Y|do(X=\hat{x}))$.}
%         \label{fig2}
% \end{figure}

% \begin{figure}[t]
%     \centering
%     \subfigure[Semeval-2016]{
%         \includegraphics[width=0.25\textwidth]{fig/fig5-2-2.png}
%     }
%     \subfigure[VAST]{
%         \includegraphics[width=0.25\textwidth]{fig/fig5-2-1.png}
%     }
%     \caption{The relationship between macro-F1 and masked ratio across different datasets. While the masked ratio is too high, we will fill in a part of random blanks at a time, iteratively until all masks are filled.}
%     \label{fig:fig_micPerMon}
% \end{figure}

\begin{figure}[t]
  \centering
  \includegraphics[width=\linewidth]{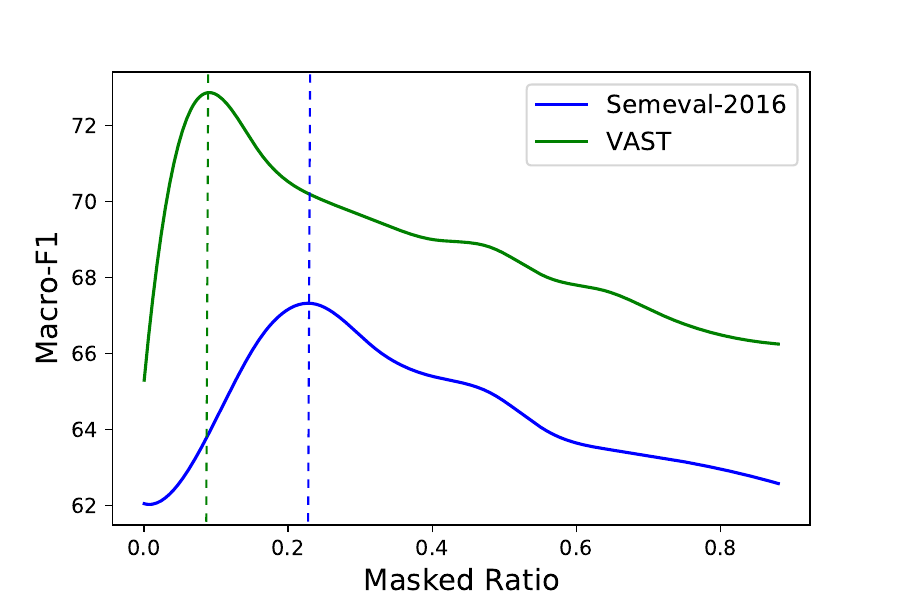}
  \caption{The relationship between macro-F1 and masked ratio across different datasets. While the masked ratio is too high, we will fill in a part of random blanks at a time, iteratively until all masks are filled.}
  \label{fig4}
\end{figure}

\begin{table*}[t]
\centering
\caption{Macro-averaged $F1$ on the VAST dataset.}
\label{tab:vast-result}
\resizebox{\textwidth}{!}{
\begin{tabular}{lcccc|cccc|cccc}
\hline
\multirow{2}{*}{Model} & \multicolumn{4}{c|}{Zero-Shot} & \multicolumn{4}{c|}{Few-Shot} & \multicolumn{4}{c}{All}       \\ \cline{2-13} 
                       & Favor  & Against & Neutral & All  & Favor & Against & Neutral & All  & Favor & Against & Neutral & All  \\ \hline \hline
Cross-Net            & .462   & .434    & .404    & .434 & .508  & .505    & .410    & .474 & .486  & .471    & .408    & .454 \\
SEKT                  & .504   & .442    & .308    & .418 & .510  & .479    & .215    & .474 & .507  & .462    & .263    & .411 \\
$\text {BERT}_\text {SEP}$            & .414   & .506    & .454    &.458 & .524  & .539    & .544    & .536 & .473  & .522    & .501    & .499 \\
$\text {BERT}_\text {MEAN}$             & .546   & .584    & .853    & .660 & .543  & .597    & .796    & .646 & .545  & .591    & .823    & .653 \\
TGA-Net                & .568   & .585    & .858    & .666 & .589  & .595    & .805    & .663 & .573  & .590    & .831    & .665 \\
CL                & .602   & .612    & .880    & .702 & \textbf{.644}  & .622    & .835    & .701 & .629  & .617    & .857    & .701 \\
$\text {C}^2\text {L}$             & .608   & .654    & .875    & .726 & .600  & .665    & .839    & .702 & .606  & .649    &\textbf{.886}    & .713 \\  \hline
RCCL(ours)                   & \textbf{.635}       & \textbf{.664}    & \textbf{.894}   & \textbf{.731}     & .637      & \textbf{.678}      & \textbf{.860}    &  \textbf{.725}    &  \textbf{.636}    & \textbf{.671}       & .877        & \textbf{.728}    \\
% w/o $h_d^-$               & .608   & .647    & .862    & .706 & .619  & .625    & .843    & .696 & .614  &  .636  & .853    & .701 \\
% w/o $h_c^-$                & .588   & .625    & .858    & .690 & .590  & .605    & .825    & .673 & .589  &  .615  & .842    & .682 \\
\hline
\end{tabular}}
\end{table*}

% \usepackage{multirow}
% \begin{table}[]
% \centering
% \caption{Macro-averaged $F1$ and variance between all targets on the SemEval-2016 dataset and UKP dataset.}
% \begin{tabular}{lccccc}
% \hline
% \multirow{2}{*}{Model} & \multicolumn{2}{c}{SemEval-2016} & \multicolumn{2}{c}{UKP} \\ \cline{2-5} 
%                        & AVG             & VAR            & AVG        & VAR        \\ \hline
% BERT                   &  .624               &                & .600           &            \\
% ~+RCCL                 &  .683               &                & .651           &            \\
% RoBERTa                &  .598               &                & .622           &            \\
% ~+RCCL                 &  .681               &                & .663           &            \\
% Flan-T5                & .710                &                & .691           &            \\
% ~+RCCL                  &  .719               &                & .706           &            \\ \hline
% \end{tabular}
% \label{tab:plms}
% \end{table}

\subsection{Hyperparameter Analysis}
\paragraph{The number of generated samples.}
Figure \ref{fig4-1} illustrates the influence of the number of positive samples and negative samples on the final results. 
Whether using generated samples for contrastive learning or data augmentation, more samples generally lead to better performance. However, the overall effectiveness of data augmentation is lower than that of contrastive learning. The analysis of experimental results suggests that the model is particularly sensitive to negative samples. When the number of negative samples is 1 and the number of positive samples is 8, the macro-F1 score is .669. However, when the number of positive samples is 1 and the number of negative samples is 8, the macro-F1 score is .651. Negative samples enhance the diversity of the original dataset due to different labels. Contrastive learning is more sensitive to negative samples than data augmentation. In contrastive learning, the objective is to make the features of all positive samples as similar as possible without negative samples. The selection of negative samples is a fundamental task in contrastive learning.

\paragraph{Masked Ratio.} 
Figure~\ref{fig4} illustrates the influence of the masked ratio on the final results. We conduct tests on two datasets: VAST, a dataset containing long texts with an average comment length of 271 characters across five sentences per sample; Semeval-2016, a dataset consisting of short texts, each sample comprising only one sentence. VAST peaks at .728 when the masked ratio is 0.08. In contrast, Semeval-2016 achieves its peak at .673 with a masked ratio of 0.2.
Datasets with larger sentence lengths should have a smaller masked ratio, and respectively, datasets with smaller sentence lengths should have a larger masked ratio. 

% \subsection{Bias Analysis}

\subsection{Few-shot and Zero-shot Experiments}
%complete
In many practical applications, training data is often too scarce. As a stress test for this data scarcity scenario, we trained our approach using the VAST dataset and evaluated it in the official test set. The overall results of our approach and baselines are shown in Table~\ref{tab:vast-result}. In order to evaluate our model properly, we respectively calculate the results on three scenarios: Zero-Shot, Few-shot, and All. Our approach achieves the best perfor in all scenarios. It illustrates the effectiveness of our approach. RCCL has a more robust performance against data scarcity. 

Among the results, CL works poorly when there is not enough data because it establishes control groups of different qualities, which limits the quality of causality estimates in data scarcity scenarios. $\text {C}^2\text {L}$ performs better in Zero-shot that Few-shot. Because data augmentation reaches a bottleneck when there is a sufficient amount of data. The growth of accuracy is limited. In the Zero-shot scenario, data is scarcer in the VAST dataset. On the other hand, our method is also a form of data augmentation. We expanded the dataset in situations of data scarcity. We enrich the labels corresponding to a certain target and complete the dataset, which is the source of our effectiveness. Then, we overcome the bottleneck of data augmentation by contrastive learning. RCCL consistently performs well regardless of the size of a given dataset, demonstrating the advantages of RCCL.

% The authors of VAST also propose five challenging phenomena. The results of the challenge are shown in Appendix.

% \subsection{Case Study}

\section{Conclusion}
In this paper, we propose a novel constrative learning framework for stance detection by mitigating pretrained stance bias to obtain ``real stance'' dependent to context features without  bias.
We firstly discovers that relative stance samples with relative bias benefit removing absolute bias and our proposed counterfactual approach can effectively measure pretrained stance bias which are challenges in stance detection.
We believe that our work could inspire researchers working on various bias caused by PLMs to develop an alternative way in removing all bias finally.
Pretrained language models hardly avoids bias including stance bias, we are interested to investigate whether our work is still available in more LLMs in future work.

\section*{Limitations}
The limitations of our current paper are as follows: (1) These results are evaluated in English; (2) The experiments are limited to BERT and its variants; and (3) The paper focuses only on stance bias. In the future, we expect these limitations above (different languages, GPT-based language models and other bias such as gender biases or sentiment biases) to be addressed. %%We error xxx

\section*{Ethics Statement}

The expressions of social comments and the stance towards targets in the paper do not represent the authors' stance. The datasets used in the paper are either open-source or used with the original authors' permission. This work specifically focuses on a targeted investigation of a particular type of bias, not encompassing all forms of bias.

\bibliography{anthology,custom}
\bibliographystyle{acl_natbib}

\appendix

\section{Appendix}
\label{sec:appendix}
\subsection{VAST Challenge}

The results of VAST challenge are shown in Table~\ref{tab:vast-challenge}. (1) Imp, the comment contains the target, and the label is non-neutral, (2) mlT, there are other examples with the same text and different targets, (3) mlS, a comment is in examples with different and non-neutral, stance labels, (4) Qte, the example contains quotes, and (5) Sarc, the example contains sarcasm. Imp examples require the model to recognize concepts related to the unmentioned target in the comment. Besides, our approach does well on mlS examples (accuracy 66.3\%). Quotes are challenging because they may repeat text with the opposite stance to what the author expresses themselves. Our approach also performs well in the Qte (accuracy 71.7\%).

\begin{table}[]
\centering
\caption{Accuracy on varying phenomena in VAST.}
\label{tab:vast-challenge}
\setlength{\tabcolsep}{1.5mm}{
\begin{tabular}{l|lllll}
\hline
Model      & Imp  & mlT  & mlS  & Qte  & Sarc \\ \hline
BERT-joint & 60.0 & 61.0 & 54.1 & 62.5 & 58.7 \\
TGA-Net    & 62.3 & 62.4 & 54.7 & 66.1 & 63.7 \\
BERT-GCN   & 61.9 & 62.7 & 54.7 & 66.8 & 67.3 \\
CKE-Net    & 62.5 & 63.4 & 55.3 & 69.5 & 68.2 \\
BS-RGCN    & 62.1 & 64.7 & 55.6 & 70.1 & 71.7\\ \hline
Ours       & \textbf{64.7} & \textbf{66.3} & \textbf{59.0} & \textbf{71.7} & \textbf{73.8} \\ \hline
\end{tabular}}
\end{table}

\end{document}